%% file: main-SIGGRAPH.tex
\documentclass[acmtog,screen,nonacm]{acmart}

\usepackage{booktabs} 

\usepackage{multirow}
\usepackage{siunitx}
\usepackage{booktabs}
\usepackage{graphicx}
\usepackage{epstopdf}
\usepackage{algorithm}
\usepackage{bm}
\usepackage{amsmath}
\usepackage{subcaption}

\usepackage[algo2e]{algorithm2e}

\SetAlFnt{\small}
\SetAlCapFnt{\small}
\SetAlCapNameFnt{\small}
\SetAlCapHSkip{0pt}

\renewcommand\footnotetextcopyrightpermission[1]{} 
\setcopyright{none}
\begin{document}
\title{Mean of Means: A 10-dollar Solution for Human Localization with Calibration-free and Unconstrained Camera Settings}

\author{Tianyi Zhang}
\email{tonax.zhang@connect.polyu.hk}
\affiliation{%
  \institution{Hong Kong Polytechnic University}
  \country{Hong Kong SAR}
}
\author{Wengyu Zhang}
\email{wengyu.zhang@connect.polyu.hk}
\affiliation{%
  \institution{Hong Kong Polytechnic University}
  \country{Hong Kong SAR}
}

\author{Xulu Zhang}
\email{compxulu.zhang@connect.polyu.hk}
\affiliation{%
  \institution{Hong Kong Polytechnic University}
  \country{Hong Kong SAR}
}

\author{Jiaxin Wu}
\email{nikki-jiaxin.wu@polyu.edu.hk}
\affiliation{%
  \institution{Hong Kong Polytechnic University}
  \country{Hong Kong SAR}
}

\author{Xiao-Yong Wei}
\email{x1wei@polyu.edu.hk }
\affiliation{%
  \institution{Sichuan University}
  \country{China}
}
\affiliation{%
  \institution{Hong Kong Polytechnic University}
  \country{Hong Kong SAR}
}
\authornote{Corresponding Author}

\author{Jiannong Cao}
\email{csjcao@comp.polyu.edu.hk}
\affiliation{%
  \institution{Hong Kong Polytechnic University}
  \country{Hong Kong SAR}
}

\author{Qing Li}
\email{qing-prof.li@polyu.edu.hk}
\affiliation{%
  \institution{Hong Kong Polytechnic University}
  \country{Hong Kong SAR}
}



\begin{abstract}
Accurate human localization is crucial for various applications, especially in the Metaverse era. 
Existing high precision solutions rely on expensive, tag-dependent hardware, while vision-based methods offer a cheaper, tag-free alternative. 
However, current vision solutions based on stereo vision face limitations due to rigid perspective transformation principles and error propagation in multi-stage SVD solvers. 
These solutions also require multiple high-resolution cameras with strict setup constraints.
To address these limitations, we propose a probabilistic approach that considers all points on the human body as observations generated by a distribution centered around the body's geometric center. 
This enables us to improve sampling significantly, increasing the number of samples for each point of interest from hundreds to billions. 
By modeling the relation between the means of the distributions of world coordinates and pixel coordinates, leveraging the Central Limit Theorem, we ensure normality and facilitate the learning process. 
Experimental results demonstrate human localization accuracy of 95\% within a 0.3m range and nearly 100\% accuracy within a 0.5m range, achieved at a low cost of only 10 USD using two web cameras with a resolution of 640×480 pixels.
%
\end{abstract}

%
%

\ccsdesc[300]{Applied computing}

\keywords{Computational geometry, point-based graphics}

\begin{teaserfigure}
    \centering
    \begin{subfigure}[b]{0.1\textwidth}
        \centering
    \end{subfigure}
    \hfill
    \begin{subfigure}[b]{0.38\textwidth}
        \centering
        \includegraphics[width=\linewidth]{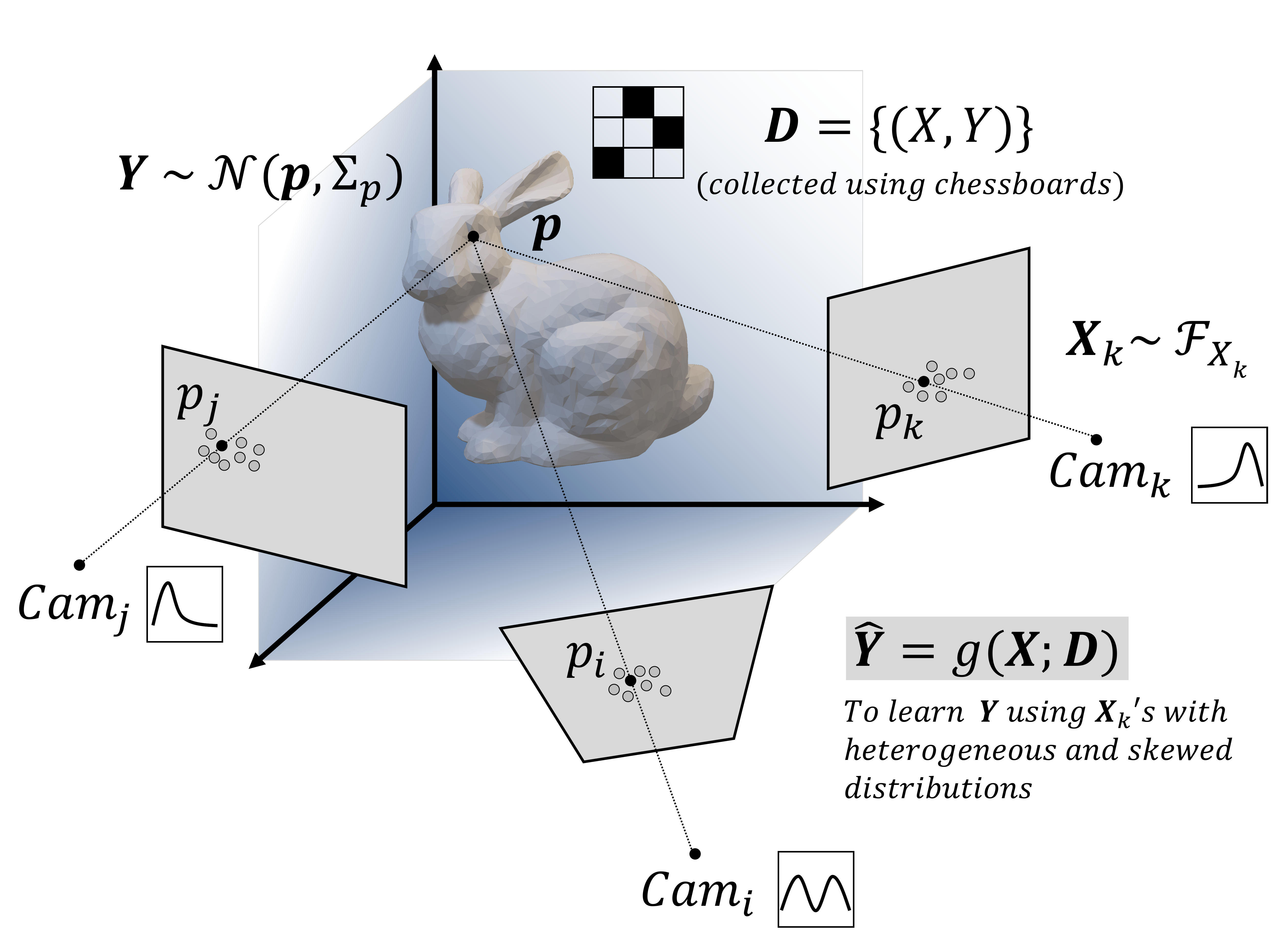}
        \caption{Traditional methods based on stereo vision}
    \end{subfigure}
    \hfill
    \begin{subfigure}[b]{0.38\textwidth}
        \centering
        \includegraphics[width=\linewidth]{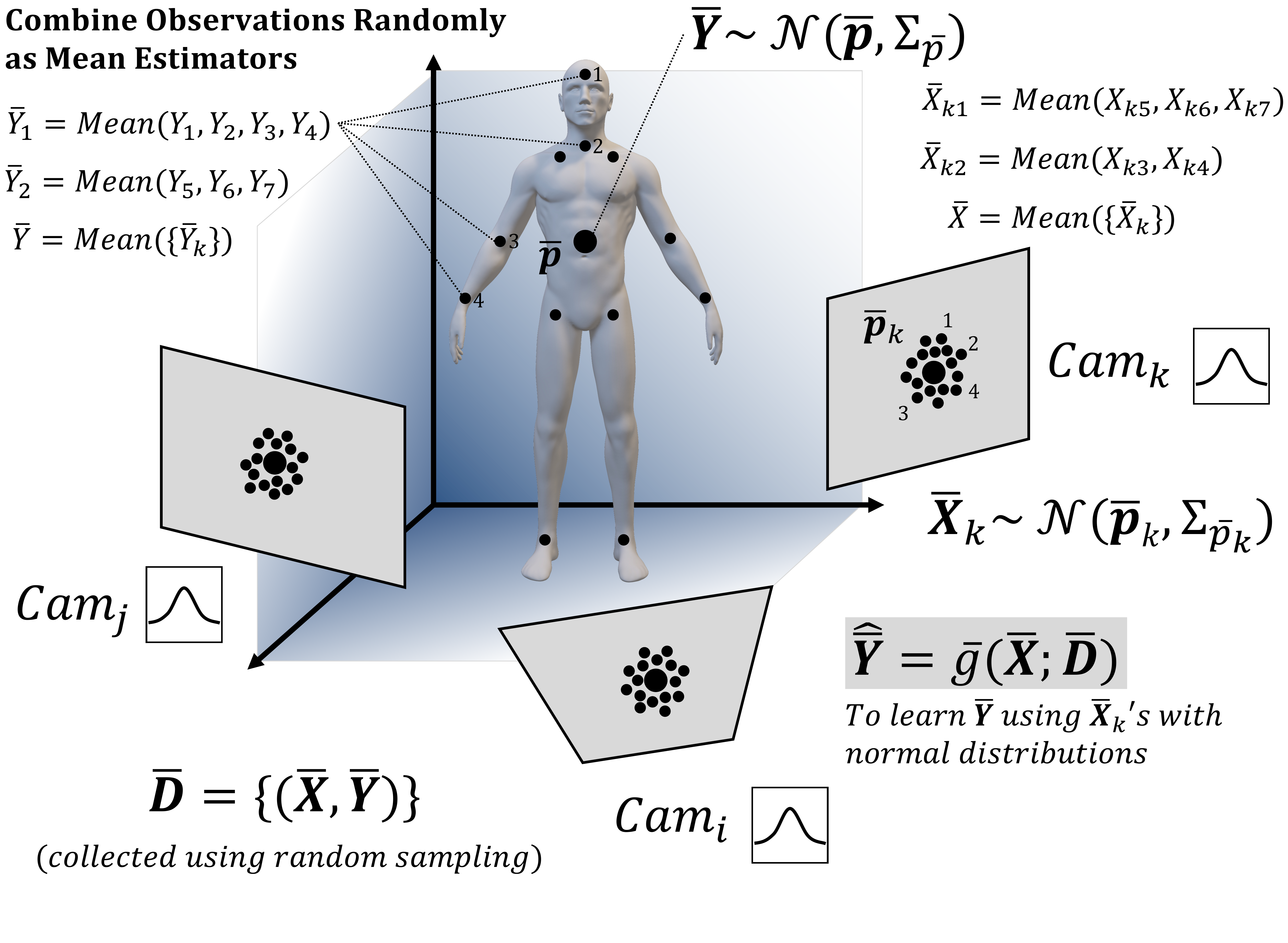}
        \caption{Mean of Means (MoM)}
    \end{subfigure}
    \hfill
    \begin{subfigure}[b]{0.01\textwidth}
        \centering
    \end{subfigure}
    \vspace{-1em}
 \caption{Comparison of traditional human localization method and Mean of Means (MoM): We have transformed the learning goal from one-to-one mapping between the world and pixel coordinates to the modeling of geometric mean distribution and the distribution of corresponding pixel coordinates.}
    \label{fig:traditional_vs_ours}
\end{teaserfigure}

\maketitle

\input{sec/1_intro}
\input{sec/2_related_work}

\input{sec/3_method}
\input{sec/5_experiment}

\input{sec/6_conclusion}

\bibliographystyle{ACM-Reference-Format}
\bibliography{reference-bibliography}

\input{sec/7_figure_only}

\end{document}


\title{Supplementary Materials: Mean of Means: A 10-dollar Solution for Human Localization with Calibration-free and Unconstrained Camera Settings}


\author{Anonymous Authors}

\maketitle

\section{Implementation details}

We use the MMPose toolbox\footnote{https://github.com/open-mmlab/mmpose} to detect 12 skeleton joint points from two input images, respectively, and then connect these joints to build a skeleton for each input image. We randomly select 30 observation points $(u,v)$ from each skeleton (i.e., $m=30$ in Eq. (\ref{eq:mean_estimator}), and taking their mean values as a $\overline{x}$. We repeat this selection 12 times, to get 12 $\overline{x}$'s for each image. 
In the training, we take the mean of 12 Kinect joint points in the world coordinate as the $\overline{y}$.  
We concatenate all $\overline{x}$ as a vector, i.e., $\overline{X}=[\overline{x}_1,...,\overline{x}_{12}]$, as the input for the MoM encoder. 
We take the mean of detected 12 skeleton joint points as the ground truth $\overline{x}$'s to compute the reconstruction loss.

The local and global MLPs in the MoM encoder has two and three residual blocks, respectively. Each residual block has three hidden layers. The dimension of all hidden layers in local MLPs is 48, while those in global MLPs is 96. The shape of the transformation matrices $\overline{P}_k$ is 3-by-4. We train the MoM model on PyTorch with a NVIDIA GeForce RTX 4090 GPU. The learning rate is $1*10^{-3}$. We use Adam optimizer. The batch size is 128. The training typically converges
after about 50 iteration. More details can be found in our code.

%% file: sec/1_intro.tex
\section{Introduction}
\label{sec:intro}

%
Human localization plays a crucial role in various applications, including AR/VR, indoor navigation, fitness and health tracking, surveillance and security, and sports analysis. 
The growing prominence of Metaverses and Digital Twins has further emphasized its significance~\cite{wang2022survey,mihai2022digital,Zhou2023MetaFi,Li2023StereoPoseAvatar}. 
For instance, in Metaverse, accurately determining user's posture and location is essential to generate a corresponding virtual representation within the Metaverse.
While posture recognition has been extensively developed~\cite{andriluka20142d,song2021human,Vladimir2021HPS,Li2023StereoPoseAvatar}, current human localization solutions primarily rely on hardware-based approaches such as UWB~\cite{poulose2021feature}, Bluetooth~\cite{li2020self}, WiFi~\cite{pu2023pacnn}, and LiDAR~\cite{dai2022hsc4d}. 
However, this hardware dependency limits their applicability in broader scenarios due to the associated costs. 
%
%
Additionally, in most hardware-based solutions, every user needs to carry a tag/device. 
Dependence on expensive tags not only adds to cost but also introduces inconvenience. 
%

In this paper, we are motivated to develop a solution using inexpensive web cameras.
By using just two cameras, costing only ten USD, we can cover a $10\times10\,m^2$ space efficiently. 
Moreover, in the context of developing applications like Metaverse, where posture recognition is typically based on computer vision techniques, it is cost-effective to leverage the existing cameras installed on-site. 
This approach also eliminates the need for users to carry additional tags, making the solution more practical.

Although it sounds promising, the use of web cameras specifically for localization has not been extensively explored in the existing literature.
While there are various vision-based tasks that resemble it, they either require strict constrained sitting or do not provide accurate world coordinates of the targets \cite{Islam2020StereoPositioning,Sinha2022TriangulationDepthStereo}. 
For instance, stereo vision methods utilize multiple cameras to analyze disparities and employ triangulation to estimate depth \cite{Sinha2022TriangulationDepthStereo}. 
However, these methods require precise camera calibration and high-resolution cameras placed in specific configurations, such as side-by-side alignment \cite{Islam2020StereoPositioning}. 
%
%
%
The expensive nature of existing solutions can be attributed to their objective of determining the depth or location of every point on the target. 
This requirement results in computationally intensive and complex solutions. 
To illustrate this concept, we present Fig.~\ref{fig:traditional_vs_ours} and examine it from a learning perspective.
Let us assume that the world coordinate of a target point $\mathbf{p}$ follows a normal distribution, resulting in observations $\mathbf{Y}\sim\mathcal{N}(\mathbf{p},\Sigma_p)$. 
The projections of $\mathbf{p}$, represented as $\{p_i\}_1^k$ on $k$ cameras, generate observations of pixel coordinates $\mathbf{X}=[\mathbf{X}_i]$. 
The training dataset, consisting of pairs of observations, is denoted as $\mathbf{D}=\{(\mathbf{X},\mathbf{Y})\}$. 
The goal of the learning process is to find a model $g$ that estimates $\mathbf{Y}$ using $\mathbf{D}$ as
\begin{equation}
    \hat{\mathbf{Y}}=g\big(\mathbf{X};\mathbf{D}\big).
    \label{eq:g_trad}
\end{equation}
This process poses challenges in three aspects.
Firstly, conventional methods require the coordinate pairs for all $\mathbf{p}$'s to cater to applications such as 3D reconstructions.
These two types of coordinates need to be aligned strictly, which makes the collecting of training data $\mathbf{D}$ difficult.
For instance, a commonly used method for camera calibration involves using chessboards, which is a tedious process of placing the boards at different angles and positions, but it can only collect samples on a limited scale of hundreds \cite{zhang2000flexible}.
Secondly, due to perspective transformation, the pixel coordinates of the target point follow unknown but skewed distributions $\mathbf{X}\sim\mathcal{F}_X$, where the pixel coordinates from different cameras actually follow different distributions $\mathbf{X}i\sim\mathcal{F}_{X_i}$. 
The heterogeneity of these distributions presents challenges for the commonly used singular value decomposition (SVD) solvers, in finding a global optimum \cite{golub1971singular, allen2016lazysvd}.
Lastly, when considering the limited size of $\mathbf{D}$ and the heterogeneity of distributions, Section~\ref{subsec:3.2} demonstrates that the expectation of the function $\mathbb{E}[\hat{\mathbf{Y}}\vert \mathbf{X};\mathbf{D}]$ does not easily converge to $\mathbb{E}[\mathbf{Y}]$.

In this paper, we present a framework called Mean of Means (MoM) to address the previously mentioned challenges. 
The fundamental idea behind is to relax the strict requirement of one-to-one pairing between world and pixel coordinates and instead consider the entire human body as a distribution generated from the mean or geometric center of the body.
The advantages are as follows.
%

\noindent\textbf{Large-Scale Sampling:} This relaxation significantly enhances the flexibility of data collection, as any sampled points from the body can be used as observations to estimate a single center. 
    Section~\ref{subsec:3.4} demonstrates that by sampling only 20 points from each body, we can theoretically collect trillions of training pairs, resulting in a considerably large training set denoted as $\overline{\mathbf{D}}$.

\noindent \textbf{Normality Consistency:} For the distribution heterogeneity, we propose modeling a relation between mean estimators rather than focusing on point estimators. 
    This leads to the formulation
    \begin{equation}
    \hat{\overline{\mathbf{Y}}}=\overline{g}\big(\overline{\mathbf{X}};\overline{\mathbf{D}}\big).
        \label{eq:g_MoM}
    \end{equation}
    As depicted in Fig.~\ref{fig:traditional_vs_ours}, this approach allows us to randomly combine observations of body points as estimators for the center $\overline{\mathbf{Y}}$. 
    While the body points initially follow a distribution determined by the body structure, the Central Limit Theorem (CLT) ensures that the sample mean will follow a normal distribution (i.e., $\overline{\mathbf{Y}}\sim\mathcal{N}(\mathbf{\overline{p}},\Sigma_{\overline{p}})$) when the sample size is sufficiently large.
    The same principle applies to the pixel coordinates, making the means of pixel coordinates also follow normal distributions (i.e., $\overline{\mathbf{X}}\sim\mathcal{N}({\mathbf{\overline{p}}_k},\Sigma_{{\overline{p}_k}})$). 
    The consistent normality of these distributions enhances the feasibility of the model learning process.

\noindent \textbf{Balance between Neural Implementation and Classical Theory:} Section~\ref{sec:method} demonstrates that the expectation of the new model $\mathbb{E}(\hat{\overline{\mathbf{Y}}})$ can converge to the true expectation $\mathbb{E}(\mathbf{Y})$, as ensured by the Law of Iterated Expectations (LIE).
    Furthermore, we propose implementing MoM using an end-to-end autoencoder framework, where an encoder learns a neural mapping function $\overline{g}(\cdot)$ directly, replacing the multi-stage SVD solvers utilized in conventional methods. 
    This approach reduces the risk of error propagation. The encoder also satisfies the conditions outlined in \cite{Alon2017GlobalOptimalCNN}, which state that a CNN with non-overlapping convolutions can achieve a global optimal solution.
    To address the potential problem of overfitting in the neural implementation of the encoder, we introduce a decoder that follows the well-established perspective transform. 
    This decoder converts the predicted $\hat{\overline{\mathbf{Y}}}$ back to its pixel coordinate correspondences $\hat{\overline{\mathbf{X}}}$ and compares them with the input $\overline{\mathbf{X}}$ to further validate the prediction.
    The collaboration between the proposed encoder and decoder ensures a balance between the capabilities of neural networks and the established perspective transformation theory.
%
Through the implementation of MoM, our experiments demonstrate that human localization accuracy can achieve 95\% accuracy within a range of 0.3m, and nearly 100\% accuracy within a range of 0.5m.
Remarkably, this level of accuracy is achieved at an on-site cost of only 10 USD, utilizing two web cameras with a resolution of 640$\times$480 pixels.

%% file: sec/2_related_work.tex
\section{Related Work}
\label{sec:related_work}
%
Human localization plays a crucial role in numerous applications, including 3D reconstruction~\cite{andriluka20142d,song2021human,Vladimir2021HPS}, healthcare~\cite{bharadwaj2017impulse,Luigi2022Healthcare}, sports analysis~\cite{ridolfi2018experimental}, video retrieval \cite{chongwahngo2005trecvid,chongwahngo2008trecvid,chongwahngo2010trecvid}.
In this section, we can only provide a concise overview of related work pertaining to our method due to space limitations. 
We have also intentionally narrowed our focus to indoor human localization within this discussion, although our proposed method holds potential for outdoor scenes.
For a more comprehensive understanding, we recommend referring to surveys in \cite{Morar2020reviewOnLocVision,Zafari2019Indoorsurvey,Yang2021SurveyIndoor,Luigi2022Healthcare}.
There are two groups of methods that have been explored: hardware-based and vision-based approaches.

\subsection{Hardware-Based Methods}
A significant portion of the work conducted in this group revolves around signal-based techniques. 
These techniques involve the installation of three or more anchor or base stations at fixed positions, which transmit and receive signals to a tag or device carried by the user.
By evaluating the time it takes for the signal to travel between the tag/device and an anchor/base station, the distance can be calculated. Subsequently, the position of the tag/device can be estimated using triangulation based on the time difference of arrival (TDOA) from all anchors/bases.
The methods within this group differ primarily in the signals utilized, such as WIFI \cite{wang2016csi,pu2023pacnn}, Bluetooth \cite{kriz2016improving,li2020self}, RFID~\cite{ruan2018device,ma2020mrliht}, and UWB \cite{vinicchayakul2014improvement,Cheng2019UWB,poulose2021feature}. 
%
%
One of the challenges faced by signal-based methods is the vulnerability of these signals to disturbances caused by the complex room environment. 
On-site objects with various layouts, materials, and surfaces can reflect signals in different patterns and significantly interfere with signal transmission and reception. 
To address this challenge, hybrid methods have been proposed, which combine different hardware components to achieve more reliable performance \cite{Monica334HybridWIFIAndUWB}.
Hardware-based methods can offer high precision, as evidenced by reported accuracies of 0.3m using UWB \cite{Wu2022HarewarePerformances}. 
However, these methods are still limited by the expensive cost of hardware, the complexity of setup, and their dependence on the specific tag/device used. 

\vspace{-1em}
\subsection{Vision-based Methods}
Vision-based methods offer a more cost-effective and straightforward setup \cite{Morar2020reviewOnLocVision}. One key advantage is their tag-free nature, making them highly applicable to a wide range of practical scenarios.
This category of methods has been extensively explored and encompasses various sub-categories. 
Due to space limitations, we will briefly introduce two categories, namely stereo vision and depth estimation.
%

%

\noindent \textbf{Stereo Vision Methods}: In stereo Vision, the theory of mapping between pixel and world coordinates has been well-established \cite{zhang2000flexible} and widely applied in various tasks. 
Typically, these methods follow a two-stage pipeline \cite{Morar2020reviewOnLocVision, wu2022indoor}.
Firstly, a Perspective-n-Point (PnP) problem is solved to determine the transformation matrix from world coordinates to pixel coordinates \cite{Min2019indoorLocalizationCameraVision,cosma2019camloc}. 
This matrix allows for the estimation of world coordinates given the pixel coordinates.
Secondly, the triangulation principle is applied, often utilizing SVD solvers, to estimate world coordinates based on the given pixel coordinates. 
It is crucial to adhere strictly to the perspective transformation principle in stereo vision theory during these calculations.
However, previous applications have demonstrated that estimating world coordinates for all points while adhering to the principle is a highly rigid requirement. 
SVD solvers can only find least square solutions for estimating all points, without ensuring consistent accuracy for each individual point. 
Additionally, this category of methods heavily relies on several assumptions, such as well-calibrated cameras \cite{yang2023Fisheye,wu2022indoor,jain2018mobiceil} and strict constraints on camera positions and angles \cite{Islam2020StereoPositioning}.
To address these limitations, bundle adjustment is frequently employed to further refine the transformation matrix and predicted world coordinates, aiming to obtain the most accurate estimations possible \cite{triggs2000bundle,zach2014robust}.
While stereo vision methods are commonly used for tasks like 3D reconstruction \cite{Johannes2016COLMAP,do2019review,Chen20123Dreconstructionstereo} and pose estimation \cite{Kang2023BinocularPose,Zhou2022QuickPoseStereoPose,Li2023StereoPoseAvatar}, our method is closely related to this category, as we apply the same theory to a different task. 
However, we have implemented our method using the mean of means approach, which relaxes the requirement of estimating all points simultaneously and eliminates the need for camera calibration and strict installation constraints.

\noindent \textbf{Depth Estimation Methods}: The primary objective of this category is to estimate depth, which refers to the distances between points on the target and the camera \cite{Zhou2017depthestimation,Bhoi2019DepthSurvey}.
Depth estimation can be achieved by leveraging motion or structural relationships among points on the target and applying principles derived from stereo vision theory to estimate scaled distances \cite{Rajasegaran2022Tracking3Dlocation,Takacs2020stereoDepthEstimate}.
In recent times, depth cameras have played a significant role in advancing depth estimation methods \cite{Bhoi2019DepthSurvey,Masoumian2022MDEreview}. 
A popular approach involves training a depth estimator using paired images and depth maps \cite{Ranftl2022RobustMonoculaDepthEstimation,Zhang_2018_DepthCompletionRGBDimage}. 
However, this approach becomes reliant on the availability of depth cameras for training, which can be both costly and susceptible to interference. 
Depth cameras typically utilize infrared technology, which shares limitations with hardware-based methods.
Moreover, the depth estimates provided by these methods do not precisely correspond to the world coordinates but rather represent scaled estimations \cite{Li2021depth2worldcoordinate}. Consequently, they are susceptible to scale ambiguity issues \cite{wang2018learning, bian2019unsupervised}.
Our method shares similarities with this category in terms of the overall task and its machine learning-based nature.

It is important to note that many methods mentioned above have extensively utilized the Microsoft Kinect, which is an RGB-D camera integrated with various vision-based techniques. 
As a result, the Kinect can be seen as a fusion of these methods and is recognized for its reliable localization accuracy, especially after scale adjustment.
However, Kinect cameras can be expensive and have limited coverage (e.g., a maximum coverage area of $5\times5\,m^2$ and perform optimally within a $3\times3\,m^2$ space).

%% file: sec/3_method.tex
\section{Method}
\label{sec:method}

\subsection{Preliminary}
Given a point $\mathbf{p}$ and its homogeneous world coordinate $\mathbf{p}_w=[\alpha,\beta,\gamma,1]^\top$, we denote its projection as a pixel (homogeneous) coordinate in the image captured by a camera as $\mathbf{p}_c=[u,v,1]^\top$.
The relation of these two coordinates is
\begin{equation}  s\,\mathbf{p}_c=\mathbf{K}\,\big[\mathbf{R}\,|\,\mathbf{T}\big]\,\mathbf{p}_w=\mathbf{P}\,\mathbf{p}_w
\label{eq:pnp}
\end{equation}
where $\mathbf{K}\in\mathbb{R}^{3\times3}$ denotes the matrix of intrinsic camera parameters, $\mathbf{R}\in\mathbb{R}^{3\times3}$ and $\mathbf{T}\in\mathbb{R}^{3\times1}$ are the extrinsic parameters defining the 3D rotation and 3D translation of the camera, and $s$ is a scale factor which is determined by the distance of $\mathbf{p}$ to the camera.
The multiplication of $\mathbf{K}$, $\mathbf{R}$, and $\mathbf{T}$ yields a transformation matrix $\mathbf{P}\in\mathbb{R}^{3\times4}$ which combines the intrinsic and extrinsic parameters.
The matrix $\mathbf{P}$ can be estimated using a set of points along with their corresponding world and pixel coordinates, known as the Perspective-n-Point (PnP) problem \cite{zheng2013revisiting,hesch2011direct}.
The Direct Linear Transformation (DLT), which replies on SVD to approximate the solution, is commonly used \cite{hartley2003multiple}.
Once $\mathbf{P}$ is estimated, it can be substituted into Eq.~(\ref{eq:pnp}) for coordinate transformation.
However, this transformation is only feasible from world coordinates to pixel coordinates (3D to 2D) because methods like DLT assume a single camera and cannot fully address the degree of freedom (DOF). 
To perform the reverse transformation (2D to 3D), multiple cameras are needed to ensure sufficient DOF. 
The Triangulation is adopted, which calculates the cross product of each side of Eq.~(\ref{eq:pnp}) using a $\mathbf{p}_c$ as
\begin{equation}
   \mathbf{p}_c\times\,s\,\mathbf{p}_c= \mathbf{p}_c\times\, \mathbf{P}\,\mathbf{p}_w,
\end{equation}
which leads to 
\begin{equation}
    \mathbf{A}\,\mathbf{p}_w=0,
    \label{eq:triangulation}
\end{equation}
where $\mathbf{A}= \mathbf{p}_c\times\, \mathbf{P}$ and the resulting $0$ is obtained because $\mathbf{p}_c\times\,\mathbf{p}_c=0$.
Eq.~(\ref{eq:triangulation}) can be utilized to estimate the $\mathbf{p}_w$ using a given set of points and their pixel coordinates in multiple cameras.
Once again, the SVD method needs to be utilized.


\subsection{Rethink From A Probabilistic  Perspective}
\label{subsec:3.2}
Let us reformulate the problem from a probabilistic standpoint. 
The objective is to predict the world coordinate (denoted by a random variable $\mathbb{E}[\mathbf{Y}]=\mathbf{p}_w\in\mathbb{R}^{4\times1}$) of a point $\mathbf{p}$ using its corresponding pixel coordinates (denoted by a random variable $\mathbb{E}[\mathbf{X}]=[\mathbf{p}_c]\in\mathbb{R}^{3\times k}$ where $k$ is the number of cameras).
This prediction is based on a training dataset $\mathbf{D}$ consisting of samples drawn from the distributions of the $\mathbf{Y}$ and $\mathbf{X}$.
As shown in Eq.~(\ref{eq:g_trad}), we can write it as a mapping function $\hat{\mathbf{Y}}=g(\mathbf{X};\mathbf{D})$.
%
The expectation of this function is
\begin{equation}
\mathbb{E}\big[\hat{\mathbf{Y}}\vert \mathbf{X};\mathbf{D}\big]=\int_{x\sim\mathbf{X}}g(\mathbf{x};\mathbf{D})\mathcal{F}_{\hat{\mathbf{Y}}\vert\mathbf{X},\mathbf{D}}(g(x)\vert x,\mathbf{D})\,dx,
    \label{eq:ex_YXD}
\end{equation}
where the $\mathcal{F}_{\hat{\mathbf{Y}}\vert\mathbf{X},\mathbf{D}}(\cdot)$ is the conditional probability density function (PDF) of $\hat{\mathbf{Y}}$ given the observation of $\mathbf{X}$ and $\mathbf{D}$.
In traditional methods using the PnP and Triangulation, the expectation of the function is not easy to converge to the expectation of $\mathbf{Y}$ (i.e., $\mathbb{E}[\mathbf{Y}]$) for three reasons.
Firstly, only a single sample $x$ can be drawn for each $\mathbf{X}$ at a specific point $\mathbf{p}$, resulting in the expectation being equivalent to
%
    $\mathbb{E}[\hat{\mathbf{Y}}\vert \mathbf{X};\mathbf{D}]=g(\mathbf{x};\mathbf{D})\mathcal{F}_{\hat{\mathbf{Y}}\vert\mathbf{X},\mathbf{D}}(g(x)\vert x,\mathbf{D})$.
%
This is considerably less efficient as a statistic for estimating the true expectation.
Secondly, the probability distribution $\mathcal{F}_{\hat{\mathbf{Y}}\vert\mathbf{X},\mathbf{D}}(\cdot)$ is likely skewed and leads to the biased predictions.
Lastly, The training dataset $\mathbf{D}$ has a limited size, typically a few hundred samples, which reduces the likelihood of the learned model $g(\mathbf{X};\mathbf{D})$ producing globally optimal solutions.
In Fig.~\ref{fig:density}, we provide a visualization of the distributions of these random variables in a real example, where the issues are evident.
%
We will design our method by addressing those issues.

\vspace{-0.1in}
\subsection{Expanding the Sampling Scope}
%
%
The issue of the sampling limitation is not easy to resolve since these methods were originally developed for tasks like 3D reconstruction where every point has to be modeled.
%
%
%
In essence, the learning process relies on a one-to-one (one $\mathbf{X}$ to one $\mathbf{Y}$) mapping, which poses challenges in sampling sufficient observations for either $\mathbf{Y}$ or $\mathbf{X}$.
%
%
In tasks related to human localization, there is no need to model every individual point. 
Instead, the target point $\mathbf{p}$ represents the geometric center of the body.
We can regard all points on the human body as the observations of the center.
Consequently, we have the opportunity to sample a large number of world (or pixel) coordinates $\mathbf{Y}_i$ (or $\mathbf{X}_i$), which are theoretically drawn from the distribution of $\mathbf{Y}$ (or $\mathbf{X}$) (see Fig.~\ref{fig:traditional_vs_ours}).
This new perspective transforms the learning process into a many-to-one problem and allows for improved design of the estimators.
For ${\mathbf{X}_i}$'s, we can gather observations by sampling points from the outcomes of skeleton detection or human detection algorithms \cite{sun2019deep}.
The detection methods have recently undergone substantial advancements, enabling us to acquire a substantial number of dependable observations.
%

\subsection{Mitigating Distribution Biases using Mean Estimators}
\label{subsec:3.4}
To mitigate the distribution biases caused by the skewness of the distributions, we design mean estimators for the $\mathbf{X}$ and $\mathbf{Y}$ as
\begin{equation}
\label{eq:mean_estimator}
    {\mathbf{\overline{X}}}=\frac{1}{m}\sum_{i=1}^m\mathbf{X}_i,\,\,{\mathbf{\overline{Y}}}=\frac{1}{m}\sum_{i=1}^m\mathbf{Y}_i
\end{equation}
where the $1\leq m\leq n$ is the number of original observations selected to construct the mean estimator and $n$ is the total number of ${\mathbf{Y}_i}$ or ${\mathbf{X}_i}$ available.
Note that we can sample multiple batches of points for both mean estimators by varying the $m$ to obtain subsets of ${\mathbf{Y}_i}$ or ${\mathbf{X}_i}$.
The maximum number of mean estimators thus becomes $\sum_{m=1}^n{n \choose m}=2^n-1$.
For example, with $20$ original observations, this can generate $1,048,575$ mean estimators.
An additional advantage is that, when building a pair $(\mathbf{\overline{X}},\mathbf{\overline{Y}})$ for learning, the mean estimators of $\mathbf{\overline{X}}$ and $\mathbf{\overline{Y}}$ is not necessarily from the same batch.
For instance, a $\mathbf{\overline{X}}$ can be calculated from pixel coordinates from the points sampled from the head, feet, and waist, which it can be paired with a $\mathbf{\overline{Y}}$ from points such as the eyes and knees.
This makes the maximum number of training pairs reach $(2^n-1)^2$, resulting in 1,099,509,530,625 pairs when $n=20$.
We denote the new training dataset of mean estimators as $\overline{\mathbf{D}}$ hereafter.
The relaxation in pairing enables a much flexible sampling which allows us to conduct any learning sufficiently.

The rational behind this flexibility is that the expectation of random variables from the same distribution is equal to that of the mean of those variables (i.e., $\mathbb{E}[\mathbf{\overline{X}}_i]=\mathbb{E}[\mathbf{\overline{X}}]=\mathbb{E}[\mathbf{{X}}]$ and $\mathbb{E}[\mathbf{\overline{Y}}_i]=\mathbb{E}[\mathbf{\overline{Y}}]=\mathbb{E}[\mathbf{{Y}}]$).
Additionally, the Law of Large Numbers (LLN) states that the average of results obtained from a large number of random samples converges to the true value as
\begin{equation}
    \mathbb{P}\Big(\lim_{n\rightarrow\infty}\mathbf{\overline{X}}={\overline{\mathbf{p}}_c}\Big),\,\mathbb{P}\Big(\lim_{n\rightarrow\infty}\mathbf{\overline{Y}}={\overline{\mathbf{p}}_w}\Big).
\end{equation}
Such a large number of samples also ensures $\mathbf{\overline{X}}$, $\mathbf{\overline{Y}}$, and $g(\mathbf{\overline{X}};\overline{\mathbf{D}})$ all follow a normal distribution as
\begin{equation}
    \mathbf{\overline{X}}\sim\mathcal{N}(\overline{\mathbf{p}}_c,\Sigma_\mathbf{\overline{X}}),\, \mathbf{\overline{Y}}\sim\mathcal{N}(\overline{\mathbf{p}}_w,\Sigma_\mathbf{\overline{Y}}),\, g(\mathbf{\overline{X}};\overline{\mathbf{D}})\sim\mathcal{N}(\overline{\mathbf{p}}_w,\Sigma_\mathbf{\overline{g}})
    \label{eq:normality}
\end{equation}
where the $\Sigma$'s are the covariance of corresponding variables.
This assurance is derived from the Central Limit Theorem (CLT), which states that the sampling distribution of the mean will always follow a normal distribution, provided that the sample size is sufficiently large. 
This helps to address the distribution bias issue in Eq.~(\ref{eq:ex_YXD}).

\subsection{Improving Prediction Accuracy using Autoencoder}
\label{sec:autoencoder}

\begin{figure}[!htb]
    \centering
    \includegraphics[width=0.48\textwidth]{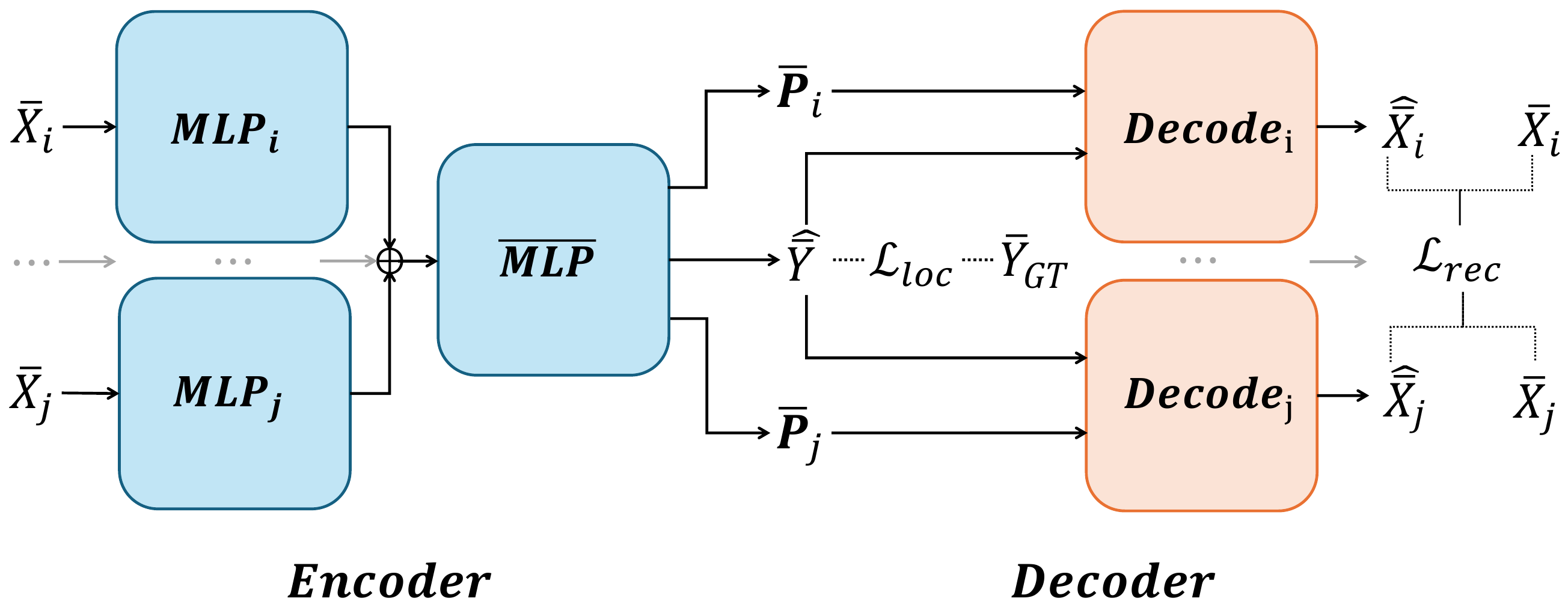}
    \caption{Neural implementation of MoM with encoder-decoder collaboration.}
    \label{fig:Network_Architecture}
    \vspace{-1em}
\end{figure}

Regarding prediction accuracy, in addition to the immediate solution of augmenting the training dataset $\mathbf{D}$ to facilitate more comprehensive learning, we propose replacing the conventional SVD solvers with an end-to-end neural network approach.
Neural networks are known to be promising function learners.
Specifically, our model follows an autoencoder structure shown in Fig.~\ref{fig:Network_Architecture}.
The encoder is a neural network implementation of the mapping function $g(\cdot)$ (Eq.~(\ref{eq:g_trad})) which predicts the mean of world coordinate from the corresponding pixel coordinates of $k$ cameras.
On the other hand, the decoder is responsible for constraining the predictions to adhere to the classical perspective model described in Eq. (\ref{eq:pnp}), thereby reducing the risk of the neural encoder learning an overfitted solution.
The design principles of our approach are as follows.

\noindent \textbf{The Mean of Means Encoder}: The encoder is a $k$-stream MLP which implements the function $g(\cdot)$ with the intention to combine the PnP and Triangulation into an end-to-end network.
As discussed early, we conduct the learning using the extensive sampling of mean pairs of $(\mathbf{\overline{X}},\mathbf{\overline{Y}})$.
As shown in Eq.~(\ref{eq:g_MoM}), the function to learn can be rewritten as $\hat{\overline{\mathbf{Y}}}=\overline{g}(\overline{\mathbf{X}};\overline{\mathbf{D}})$.
%
%
We refer to this model as Mean of Means (MoM) since it predicts the mean of the world coordinates using mean estimators of pixel coordinates. 
MoM enhances prediction accuracy from three perspectives.

Firstly, this end-to-end architecture reduces the risk of error propagation in traditional multi-stage SVD solvers and increases the likelihood of obtaining a globally optimal model.
Each stream of the MLP corresponds to a specific camera.
It takes from each camera a set of mean estimations and feed them into a local MLP (denoted as $\textbf{MLP}_k$) for camera-specific parameter learning.
The features of all streams are then concatenated and fed into a global MLP (denoted as $\overline{\textbf{MLP}}$) for world coordinate prediction.
The neural implemented of the encoder is then written
%
\begin{equation}
    \overline{g}(\mathbf{\overline{X}};\overline{\mathbf{D}})=\overline{\textbf{MLP}}\Bigg(\bigoplus_k\textbf{MLP}_k\Big(\mathbf{\overline{X}}_k\Big)\Bigg),
\end{equation}
where $\oplus$ is for feature concatenation.
Note we set the size of mean estimator set to 12 (i.e., $\mathbf{\overline{X}}_k\in \mathbb{R}^{12\times2}$) in the sense that in traditional PnP, there are 12 DOF.
The MLPs consist of fully connected layers (FCs) and use ReLU as the activation function.
%
%
To conduct the learning for the encoder, we employ L2 norm as the loss function for the encoder
\begin{equation}
\mathcal{L}_{loc}=\sum_{\overline{x}\sim\mathbf{\overline{X}}}\Big\|\,\overline{g}(\mathbf{\overline{x}};\overline{\mathbf{D}})-\overline{y}\,\Big\|,
\label{eq:L_loc}
\end{equation}
where every training pair $(\overline{x},\overline{y})\in \overline{\mathbf{D}}$ is sampled from the possible mean estimator pairs $(\mathbf{\overline{X}},\mathbf{\overline{Y}})$.

The design of this encoder is inspired by the findings in \cite{Alon2017GlobalOptimalCNN}, where the authors discovered that a CNN with non-overlapping convolutions and ReLU activations can guarantee a globally optimized solution when the inputs follow a normal distribution.
Our model consists of FC layers, which can be viewed as a special CNN with $1\times1$ non-overlapping convolutions. 
Moreover, our inputs, i.e., the mean estimators, follow a normal distribution. 
Therefore, the model satisfies the required conditions and is expected to generate predictions with higher accuracy.

Secondly, our implementation of predicating the mean using mean estimators has indeed transferred the problem of predicting the single-point coordinate into a problem of estimating the expectation of the coordinate.
Furthermore, our extensive sampling ensures we have a comprehensive coverage of the supports of the $\mathbf{\overline{X}}$, $\mathbf{\overline{Y}}$, and $\mathbf{\overline{D}}$.
By the Law of Iterated Expectations (LIE), the expectation of the function $\mathbb{E}(\overline{g}(\mathbf{\overline{X}};\overline{\mathbf{D}}))$ 
converges to the $\mathbb{E}(\mathbf{{Y}})$.
This can be proven as 
\begin{align}
    \mathbb{E}\Big(\overline{g}\big(\mathbf{\overline{X}};\overline{\mathbf{D}}\big)\Big)&= \mathbb{E}\Big(\mathbb{E}\big(\hat{\mathbf{\overline{Y}}}\vert \mathbf{\overline{X}},\overline{\mathbf{D}}\big)\Big)\nonumber\\
   &=\int_{\overline{x}\sim\overline{X}}\int_{\overline{d}\sim\overline{D}}\mathbb{E}\Big(\hat{\mathbf{\overline{Y}}}\vert \mathbf{\overline{x}},\overline{\mathbf{d}}\Big)\mathbb{P}\big(\mathbf{\overline{x}},\overline{\mathbf{d}}\big)\,d\overline{d}\,d\overline{x}\,\,(\text{by LOTUS})\nonumber\\
   &=\int_{\overline{x}\sim\overline{X}}\int_{\overline{d}\sim\overline{D}}\Bigg(\int_{\hat{\overline{y}}\sim\hat{\overline{Y}}}\hat{\overline{y}}\,\mathbb{P}\big(\hat{\mathbf{\overline{y}}}\vert \mathbf{\overline{x}},\overline{\mathbf{d}}\big)\,d\hat{\overline{y}}\Bigg)\mathbb{P}\big(\mathbf{\overline{x}},\overline{\mathbf{d}}\big)\,d\overline{d}\,d\overline{x}\nonumber\\
   &=\int_{\overline{x}\sim\overline{X}}\int_{\overline{d}\sim\overline{D}}\int_{\hat{\overline{y}}\sim\hat{\overline{Y}}}\hat{\overline{y}}\,\mathbb{P}\big(\hat{\mathbf{\overline{y}}}\big)\,d\hat{\overline{y}}\,d\overline{d}\,d\overline{x}\nonumber\\
   &=\mathbb{E}\big(\hat{\mathbf{\overline{Y}}}\big)=\,\mathbb{E}\big(\overline{\mathbf{Y}}\big)=\,\mathbb{E}\big(\mathbf{Y}\big),\,\,(\text{by Eq.~(\ref{eq:normality}) and (\ref{eq:L_loc}) })
\end{align}
where the LOTUS stands for the Law of the Unconscious Statistician.

Lastly, the use of mean estimators and the assurance of normality not only increase the flexibility of sampling and learning, but also enhance the model's resistance to noise and perturbations by allowing proper cancellation of positive and negative biases.
In our experiments, we will demonstrate the robustness of the model against camera motions and various types of noise.

\noindent \textbf{Perspective Transformation Regulated Decoder}: Neural networks show promise as function learners.
However, they often face the issue of overfitting, especially when dealing with a large number of parameters compared to traditional methods (e.g., only 12 parameters in the transformation matrix $\mathbf{P}$).
To address this, we introduce a decoder that works in tandem with the MoM encoder.
The concept is that if the encoder accurately predicts a world coordinate $\mathbf{\overline{Y}}$ for a point $\mathbf{p}$, then the predicted $\hat{\mathbf{\overline{Y}}}$ should be convertible back to $\mathbf{p}$'s pixel coordinate $\mathbf{\overline{X}}$ using a traditional perspective transformation (e.g., Eq.~(\ref{eq:pnp})). 
Since the theory behind traditional perspective transformation is well-established, the decoding process encourages the encoding learner to avoid learning an overly complex function. 
This enables validation of predictions and regulates the learning process.

To achieve this, we extend the MoM encoder's output with $k$ transformation matrices ${\overline{\mathbf{P}}_k}$. 
It is important to note that these ${\overline{\mathbf{P}}_k}$ serve as analogies for $\mathbf{P}$ in Eq.~(\ref{eq:pnp}). 
In traditional SVD solvers, the factor $s$ is cancelled during the calculation, which is one reason that accurate world coordinates cannot be estimated. 
In an end-to-end neural implementation, we can overcome this limitation and learn the factor $s$ along with $\mathbf{P}$.
Each transformation matrix $\overline{\mathbf{P}}_k$ is essentially an estimation of $\frac{\mathbf{P}}{s}$.
The decoder for regulation is then a $k$-stream MLP where in each stream, the calculation is equivalent to traditional perspective transformation of 
\begin{equation}
\hat{\mathbf{\overline{X}}}=\textbf{Decode}_k\Big(\hat{\mathbf{\overline{Y}}}\Big)=\overline{\mathbf{P}}_k\hat{\mathbf{\overline{Y}}}.
\end{equation}
The loss for the decoder is written
\begin{equation}
\mathcal{L}_{rec}=\sum_{k}\Big\|\,\textbf{Decode}_k\Big(\hat{\mathbf{\overline{Y}}}\Big)-\mathbf{\overline{X}}_k\,\Big\|,
\end{equation}
where $\mathbf{\overline{X}}_k$ is the mean of pixel observations of the $k^{th}$ camera.
The encoder and decoder work together to form an autoencoder, collectively addressing the learning issues.

%% file: sec/5_experiment.tex
\begin{table*}[]
\caption{Performance comparison on human localization with the traditional method (PnP+Triangulation) and UWB. The best results are in bold font.}
\vspace{-1em}
\label{tab:comparison_with_other_methods}
\resizebox{\linewidth}{!}{

\begin{tabular}{c|c|ccc|cc|cccc}
\toprule
      \multirow{2}{*}{Walk Patterns}  &\multirow{2}{*}{Methods}       & \multicolumn{3}{c|}{Position Error (m)} & \multicolumn{2}{c|}{Trajectory Error (m)} & \multicolumn{4}{c}{Accuracy at Different Thresholds (\%)} \\
          &    & Mean $\downarrow$   & Median $\downarrow$     & Std. $\downarrow$     &   ATE $\downarrow$                          & RPE $\downarrow$                             & @0.2m $\uparrow$      & @0.3m $\uparrow$      & @0.4m $\uparrow$     & @0.5m $\uparrow$     \\\hline

\multirow{3}{*}{Random Walk} &PnP + Triangulation & 0.426	&0.341	&0.333	&0.705	&0.115 &21.960	&41.718	&	59.449		&71.027       \\
&UWB \cite{Cheng2019UWB}    & 0.263&	0.258&	0.120	&0.360&	0.334	&33.246	&61.753&	86.016&	97.507
 \\
&MoM (ours)    & \textbf{0.150}	&\textbf{0.132}	&\textbf{0.114}	&\textbf{0.279}	&\textbf{0.087}	&\textbf{80.111}	&\textbf{95.115}	&\textbf{98.542}	&\textbf{99.213}     			       \\\hline

\multirow{3}{*}{Cross Walk}&PnP + Triangulation  & 0.473	&0.328	&0.472	&0.732	&0.143	&28.905	&45.156		&63.503		&74.654   \\
&  UWB \cite{Cheng2019UWB}   & 0.266 &	0.263	&0.117	&0.281&	0.335&	30.407&	61.408	&85.646	&97.983
 \\
&MoM (ours)    &\textbf{0.158}	&\textbf{0.134}	&\textbf{0.118}	&\textbf{0.286}	&\textbf{0.101}	&\textbf{78.529}	&\textbf{93.278}	&\textbf{97.034}	&\textbf{98.458}   			       \\ \hline

\multirow{3}{*}{Square Walk}&PnP + Triangulation  &0.642	&0.554	&0.414	&0.879	&0.128&	10.130	&20.975&	35.267	&45.986  \\    
&   UWB \cite{Cheng2019UWB}   & 0.267&	0.262	&0.117	&0.294	&0.327&	31.358&	61.496&	85.288	&97.604
 \\
& MoM (ours)   & \textbf{0.142}	&\textbf{0.126}	&\textbf{0.076}	&\textbf{0.244}	&\textbf{0.082}	&\textbf{81.883}	&\textbf{95.544}	&\textbf{99.285}	&\textbf{100.000} \\ \hline

\multirow{3}{*}{Overall}&PnP + Triangulation  &0.514	&0.408	&0.407	&0.772	&0.129	&20.332 	&35.950 	&52.740 	&63.889  \\    
&   UWB \cite{Cheng2019UWB}  &0.265 	&0.261 &	0.118 &	0.312 	&0.332 &	31.670 &	61.552&	85.650 &	97.698 \\
& MoM (ours)  & \textbf{0.150} &	\textbf{0.131} &	\textbf{0.103} &	\textbf{0.270} &	\textbf{0.090} &	\textbf{80.174} &	\textbf{94.646} 	&\textbf{98.287} 	&\textbf{99.224} 
      \\
\bottomrule
\end{tabular}
}
\end{table*}

\section{Experiment}
\label{sec:experiment}
%

We implement MoM using PyTorch and build a benchmark dataset for experiments.
All experiments were conducted on a system equipped with 64 CPU cores, 96 GB RAM, and an NVIDIA RTX 4090 GPU.
%

%
\vspace{-1em}
\subsection{Benchmark Dataset}
\label{sec:dataset}
To evaluate the proposed method, we construct a human localization benchmark in a real-world indoor environment. 
The dataset consists of over 34k samples of human location in both world and pixel coordinates. 
We collect human walking data in an indoor space of 10 $\times$ 10 $m^2$. 
We use two web cameras (MF-100) to record videos (640 $\times$ 480 pixels) of humans walking at two different angles for pixel coordinates. 
Besides, we also use a Kinect (Azure Kinect DK) as well as a UWB system (BP-TWR-50) to capture world coordinates. 
The layout and the parameters of the web cameras, Kinect, and UWB can be found at our GitHub depository. 
Fifteen subjects (seven males and eight females) participate in the data collection. 
The participants were instructed to perform three walking tasks that encompassed three distinct walking patterns: random walking within the space, walking along a predefined path with a cross shape (cross walk), and walking along a predefined square path measuring (square walk). 
For the random walk, each participant is required to walk for around 5 minutes. 
For the cross and square walks, each participant needs to step in the predefined paths two times. 
The ground truth of the human location in the world coordinate is obtained by taking the mean of real-world locations of 12 Kinect joint points. 
Specifically, these 12 points are left and right joint points of shoulders, elbows, wrists, hips, knees, and ankles. 
%

The data was divided into training and testing sets based on participants and walking patterns. 
For most experiments in this section, unless otherwise specified, the training set consisted of walking data from 10 randomly selected individuals, while the testing set comprised data from the remaining 5 individuals. 
In Section~\ref{subsec:4.5}, we explore different approaches to splitting the datasets.
Furthermore, during training, only random walk data was used to prevent the model from learning specific walking patterns. 
However, all walking patterns were employed during testing. 
Consequently, the training set consisted of 13,950 walk samples, while the testing set included 6,858, 2,529, and 2,379 samples of random walk, cross walk, and square walk, respectively.




\subsection{Evaluation Metrics}
We use several metrics to evaluate the performance of human localization, including 1) Positioning Errors: we report the Mean, Median and Standard deviation;
2) Localization Accuracy: we use several thresholds, i.e., Acc@0.2m, Acc@0.3m, Acc@0.4m, Acc@0.5m, where Acc@0.2m is the accuracy that a prediction is counted as accurate if the distance between the predicted location and the ground truth is less than 0.2 meters;
and 3) Trajectory Errors: we report Absolute Trajectory Error (ATE) and Relative Pose Error (RPE).

\subsection{Comparison with Other Methods}

We compare our proposed model MoM with the traditional PnP and Triangulation solution and a hardware-based method (i.e., UWB) \cite{Cheng2019UWB}. 
%
%
Although PnP and Triangulation has been widely used in 3D reconstruction \cite{Johannes2016COLMAP}, there are a few studies using it for human localization. 
We use the PnP and Triangulation packages on OpenCV to implement Eq. (\ref{eq:pnp}) and Eq. (\ref{eq:triangulation}) as a traditional solution for human localization. 
Besides, the camera is calibrated using \cite{zhang2000flexible} before applying PnP to calculate the extrinsic parameters. 
We also compare MoM with UWB method \cite{Cheng2019UWB}. 
UWB uses high-bandwidth radio communications, which is considered an effective and representative indoor human localization approach with promising accuracy. 

Table~\ref{tab:comparison_with_other_methods} compares their performances on the test sets of three walk patterns. Our method MoM consistently outperforms the traditional PnP + Triangulation and UWB on all evaluation metrics over three walk patterns. 
Specifically, our method obtains a mean positioning error of $0.150m$. 
In contrast, the corresponding errors of PnP + Triangulation and UWB are $0.514m$ and $0.265m$.
Our method also has the lowest trajectory error (0.270 and 0.090 on ATE and RPE, respectively). 
Moreover, in terms of location accuracy, our method achieves Acc@0.3m of over 90\%. It is important to note that the UWB devices used in our experiments are development kits primarily intended for research purposes. These devices may have lower accuracy compared to professional ones on the market.

We also compared the efficiency of predicting human locations. Out of the 6,858 samples, the speed of the conventional method, UWB, and our MoM were 0.24 seconds, 745.44 seconds, and 6.47 seconds, respectively, 
on CPU. 
It is noteworthy that our MoM model achieves an inference rate of 1,060 samples per second when using observation points as input. This demonstrates the real-time capability of our method.

\subsection{Collaboration of the Encoder and the Decoder}

\begin{figure}[t]
    \centering
    \begin{subfigure}[b]{0.49\linewidth}
        \centering
        \includegraphics[width=\linewidth]{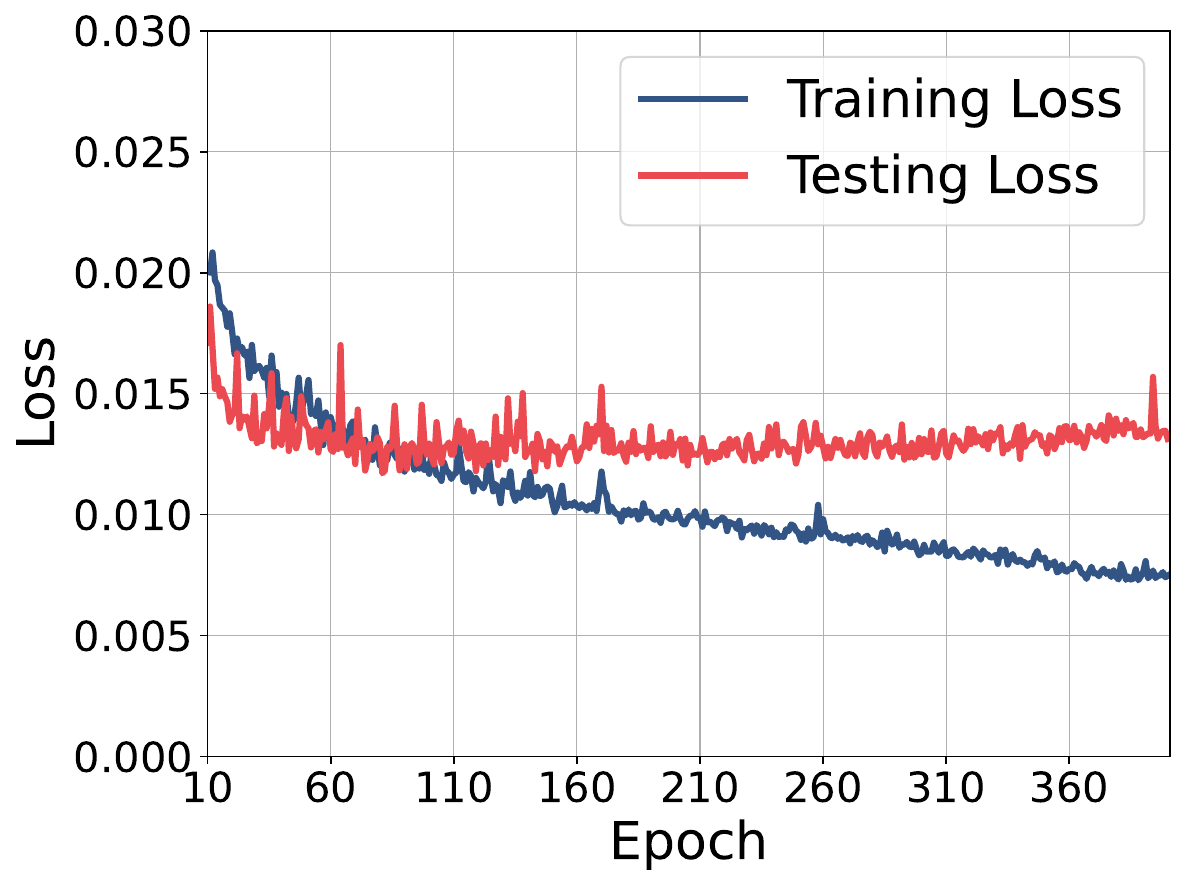}
        \vspace{-1em}
        \caption{With decoder.}
    \end{subfigure}
    \hfill
    \begin{subfigure}[b]{0.49\linewidth}
        \centering
        \includegraphics[width=\linewidth]{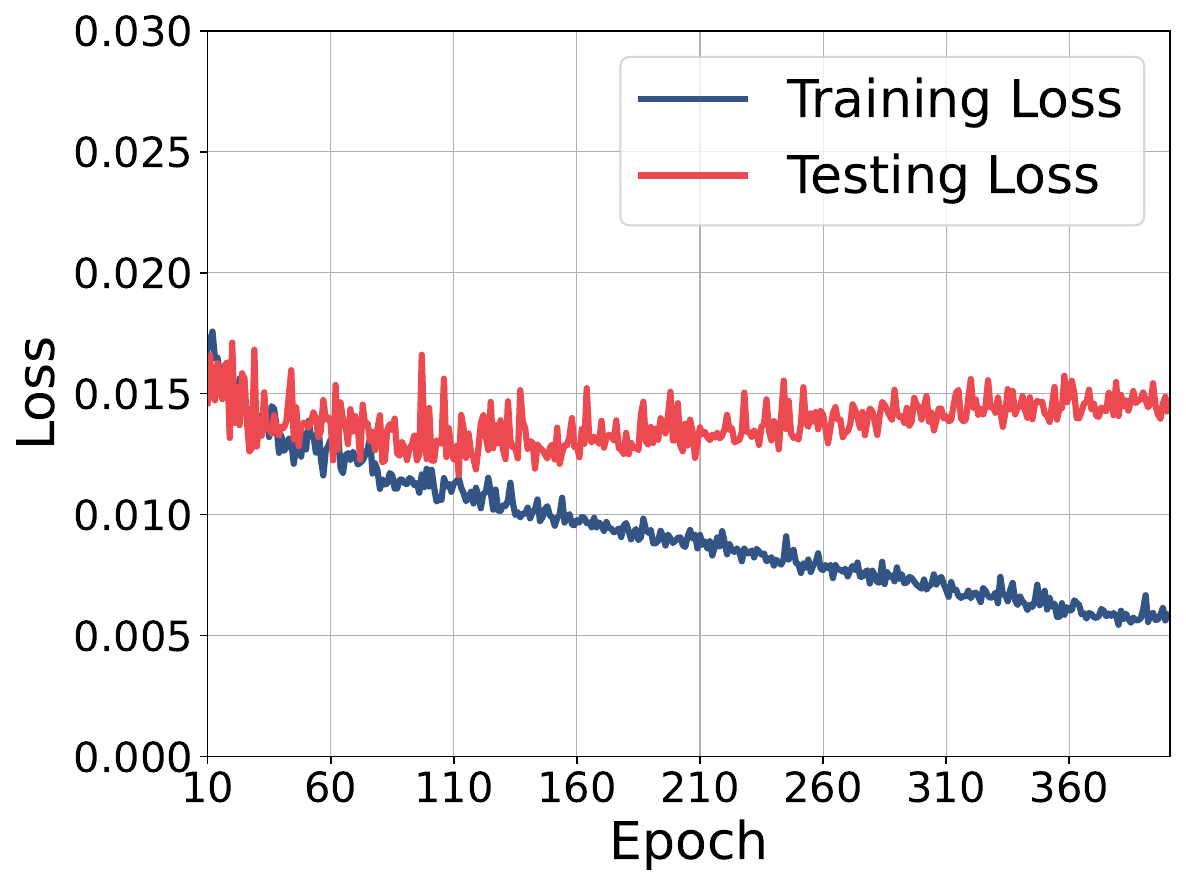}
        \vspace{-1em}
        \caption{Without decoder.}
    \end{subfigure}
    \hfill
    \vspace{-1em}
    \caption{Training and testing losses of MoM.}
    \label{fig:loss_with_and_without_decoder}
    \vspace{-1em}
\end{figure}

To evaluate the collaboration of the encoder and decoder in MoM for overfitting issues (Section~\ref{sec:autoencoder}), we compare the performances of the MoM model with and without the decoder in Table~\ref{tab:effect_of_encoder}. 
With the help of the decoder, the encoder achieves higher accuracy at almost all thresholds. 
The effectiveness of the collaboration of the encoder and decoder is further verified by the training and testing losses in Fig.~\ref{fig:loss_with_and_without_decoder}. 
With the decoder, the testing loss is stable and is around 0.013 after 60 epochs. 
However, without the decoder as a regulation, the testing loss increases as the MoM network is trained with more epochs (e.g., 0.015 at the epoch of 400). 
Moreover, with the decoder, the testing loss has a lower variance than that without the decoder.

\begin{table}[]
\caption{Performance of MoM with and without the decoder regulation.}
\vspace{-1em}
\label{tab:effect_of_encoder}
\resizebox{\linewidth}{!}{

\begin{tabular}{c|c|cccc}
\toprule
     \multirow{2}{*}{Walk Patterns}  & \multirow{2}{*}{Setting}    &  \multicolumn{4}{c}{Accuracy at Different Thresholds (\%)} \\
       &&    @0.2m $\uparrow$      & @0.3m $\uparrow$      & @0.4m $\uparrow$     & @0.5m $\uparrow$     \\\hline

\multirow{2}{*}{Random Walk} &w/o Decoder &78.944	&94.605	&98.527	&\textbf{99.315}\\
                             &w/ Decoder & \textbf{80.111}	&\textbf{95.115}	&\textbf{98.542}	&99.213\\ \hline
\multirow{2}{*}{Cross Walk} &w/o Decoder &74.852			&\textbf{93.673}	&\textbf{97.034}	&98.300\\
                             &w/ Decoder & \textbf{78.529}	&93.278	&\textbf{97.034}	&\textbf{98.458}\\\hline
\multirow{2}{*}{Square Walk} &w/o Decoder &77.512		&94.283	 &98.571	&99.538\\
                             &w/ Decoder &\textbf{81.883}	&\textbf{95.544}	&\textbf{99.285} &\textbf{100.000}\\ \hline
\multirow{2}{*}{Overall} &w/o Decoder &77.103 	&94.187 &	98.044 &	99.051 \\ 
                         &w/ Decoder &\textbf{80.174} &	\textbf{94.646} 	&\textbf{98.287} 	&\textbf{99.224} \\
\bottomrule

\end{tabular}
}

\end{table}

\subsection{Performance Over Number of Participants in Training}
\label{subsec:4.5}
The robustness of the MoM model is evaluated across different participants in our study. 
We specifically assess the human localization performance by varying the number of participants included in the training set. 
For each run, we randomly select $n$ out of 10 participants to form the training set. 
This selection process is repeated three times at each $n$, and the average accuracy is reported.

Fig.~\ref{fig:diff_num_people} illustrates the comparison of localization accuracy across different values of $n$ for all testing samples. Notably, even with training data from a single individual, our method achieves an accuracy exceeding 90\% Acc@0.4m. 
As the number of participants increases to four, the performance stabilizes and approaches the best accuracy, except for Acc@0.2m. 
Additionally, when more than three people are included in the training set, our model exhibits reduced sensitivity to human selections, as evidenced by a standard deviation of less than 1\% for Acc@0.3m, 0.4m, and 0.5m. 
These results validate the robustness of the MoM model concerning the number of participants and its independence from specific walking patterns. 
In other words, when predicting the location of a new user, the MoM model does not require fine-tuning specifically for that individual.

\subsection{Performance Over Camera Motions}
We verify the effectiveness of our method against different degrees of camera perturbation. 
Specifically, we stimulate the perturbation by adding random pixel offsets to the two input images and then sample the observation points from the perturbed images. 
Fig.~\ref{fig:camera_jitter} compares the performance with a pixel offset from 0 to 19. With an offset less than 8, there are no significant changes on Acc@0.3m. With an offset of 10, the Acc@0.3m has a slight drop by 3\%.

\subsection{Performance Over Noise}
We also verify the robustness of our MoM method over different degrees of Gaussian noise. 
Specifically, we add a 2D Gaussian noise with zero mean and std equal to $n$ pixels to the detected skeleton joint points. 
Then, the detected joint points are shifted to a random point of the Gaussian distribution to build up a new noisy skeleton. 
The observation points are subsequently sampled from the new skeleton for location prediction. 
Fig.~\ref{fig:noisy_sampling} compares the performance with $n$ ranging from 0 to 19. There are no significant performance differences on Acc@0.4m and 0.5m with different noise values. When $n=19$, Acc@0.3m has a slight drop by 3\%. 


\subsection{Applications}
In order to showcase the practicality of the MoM approach in various applications, we have successfully integrated it with two avatars. 
The first avatar is a walking robot implemented in Unity, while the second avatar is Steve from Minecraft. 
Several examples of these applications can be seen in Fig.~\ref{fig:application}.
For additional videos and more examples, please refer to the supplementary materials.

%% file: sec/6_conclusion.tex
\section{Conclusion}
\label{sec:conclusion}
This paper introduces a unified framework called Mean of Means (MoM) to address challenges related to one-to-one pairing of world and pixel coordinates in human body localization. 
MoM relaxes this requirement by considering the entire body as a distribution generated from the mean or geometric center. 
The advantages of MoM include large-scale sampling, normality consistency, and a balance between neural implementation and classical theory. Large-scale sampling allows flexibility in data collection, enabling the estimation of a single center using sampled points from the body. 
Normality consistency models the relation between mean estimators, ensuring that the means of body and pixel coordinates follow normal distributions. 
The balance between neural implementation and classical theory is achieved through an end-to-end autoencoder framework, replacing multi-stage solvers with neural mapping functions and leveraging a decoder for validation. The performance of MoM has been validated in the experiments.

%% file: sec/7_figure_only.tex
\begin{figure*}[t]
    \centering
    \includegraphics[width=0.95\textwidth]{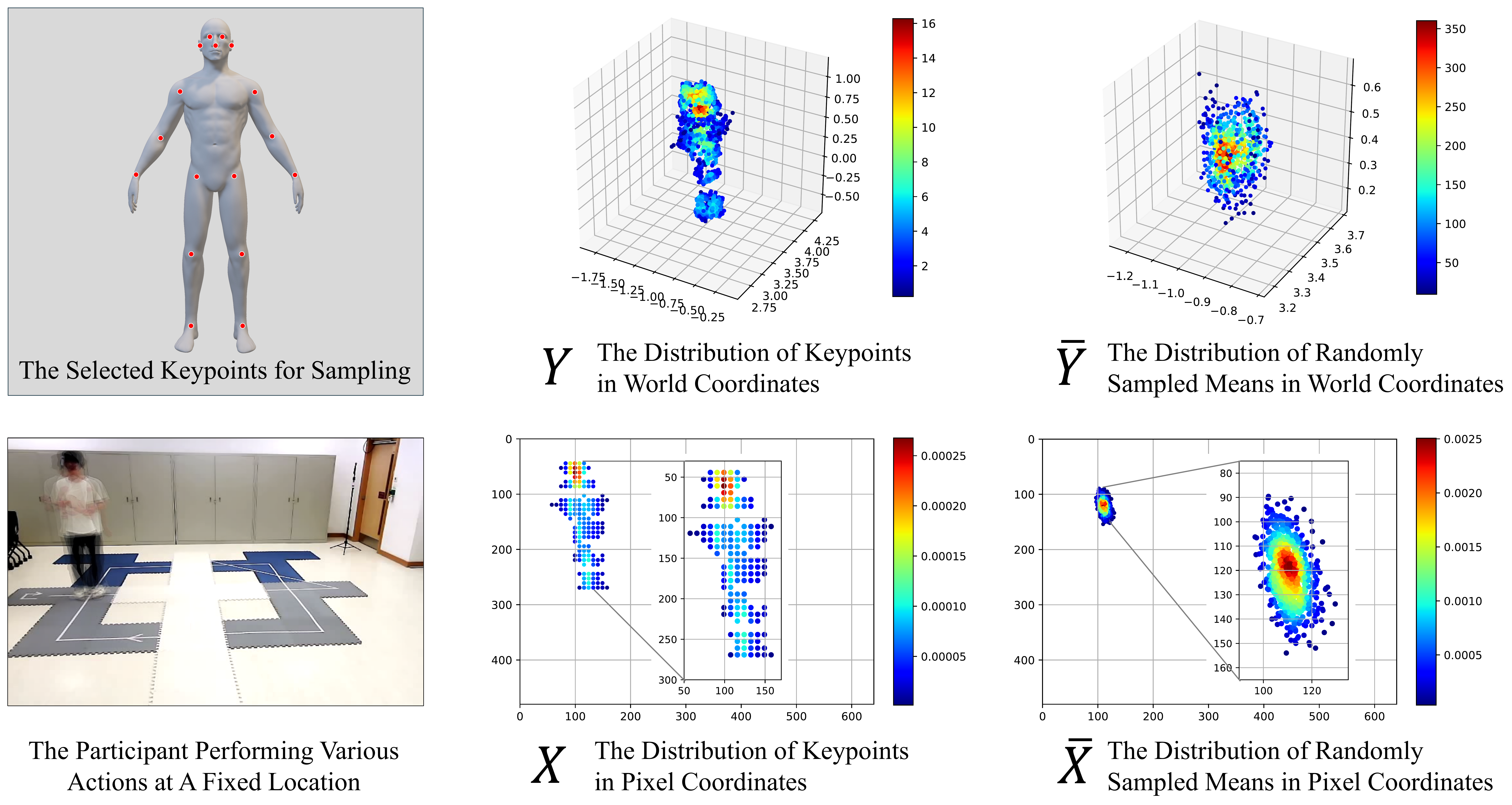}
    \caption{Distributions of different random variables in a real example: the camera captures the user standing in a fixed location and performing various actions repeatedly. The world and pixel coordinates of the keypoints on the body at different time points can then be used as the observations of variables.}
    \label{fig:density}
\end{figure*}

\begin{figure*}[t]
    \centering
    \includegraphics[width=1\textwidth]{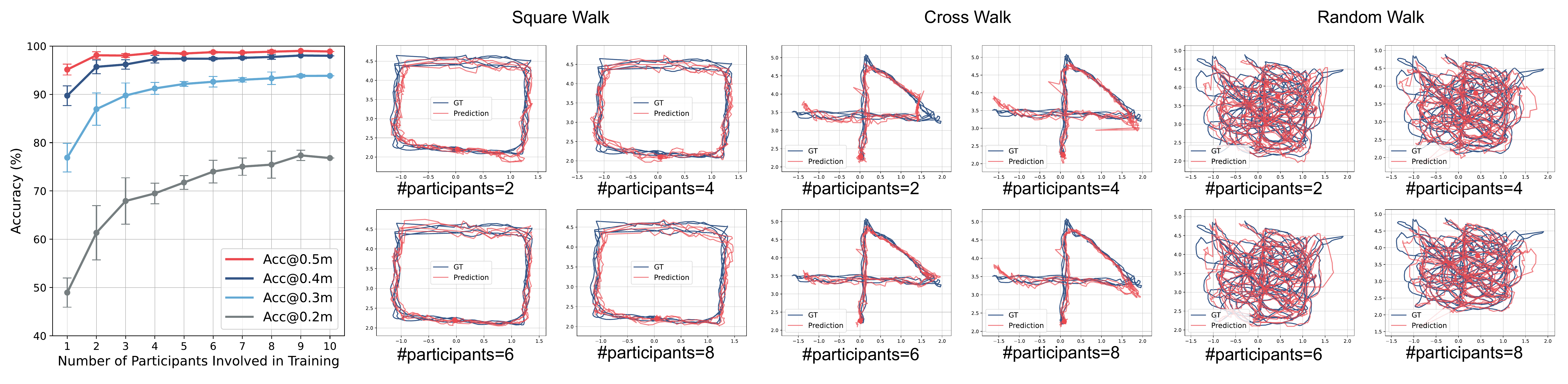}
    \caption{The MoM performance over number of participants involved in training with the predicted trajectories compared with the ground truth (GT): as the number of participants increases, the model's performance evolves and stabilizes. The performance appears to reach a convergence point when using four participant's data for training. }
    \label{fig:diff_num_people}
\end{figure*}

\begin{figure*}[t]
    \centering
    \includegraphics[width=1\textwidth]{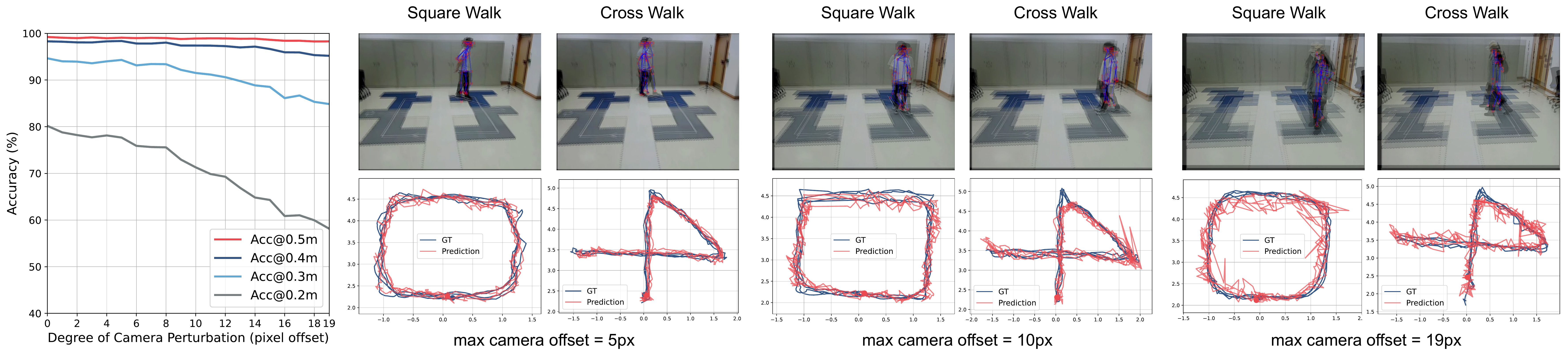}
    
    \caption{The performance of the MoM method was evaluated under varying degrees of camera perturbation with the predicted trajectories compared with the ground truth (GT): no significant drops in performance within the range of 0.3m were observed until the maximum camera offsets exceeded 8 pixels.}
    \label{fig:camera_jitter}
\end{figure*}

\begin{figure*}[t]
    \centering
    \includegraphics[width=1\textwidth]{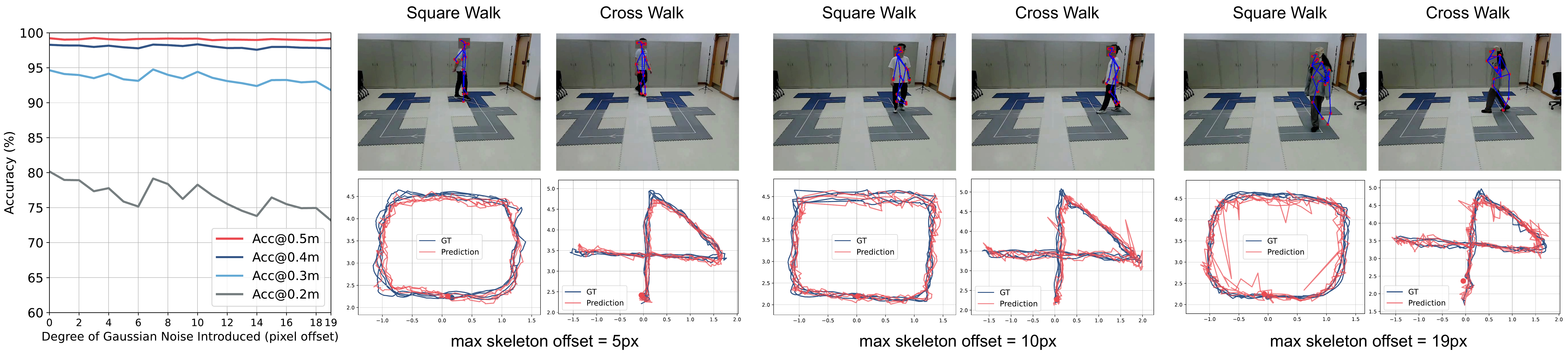}
    \caption{The performance of the MoM method over various degrees of noise introduced into the results of skeleton detection: the keypoints on the skeleton have been deteriorated with offsets in pixels. No significant drops in performance were observed within all ranges.}
    \label{fig:noisy_sampling}
\end{figure*}


\begin{figure*}[t]
    \centering
    \includegraphics[width=1\textwidth]{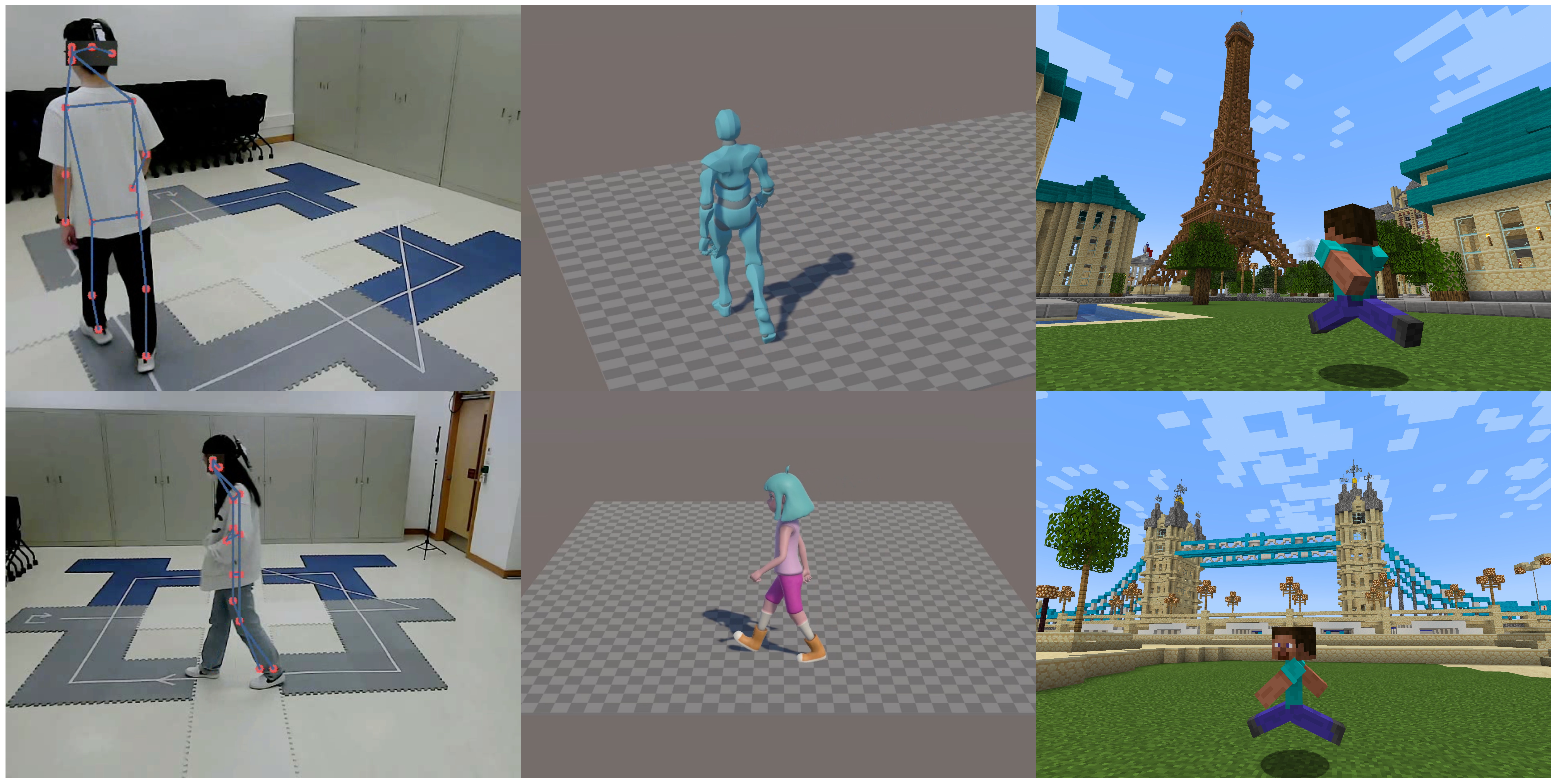}
    \caption{Applications by integrating MoM with Avatars in Unity and Minecraft.}
    \label{fig:application}
\end{figure*}